\documentclass[a4paper]{article}

\usepackage{INTERSPEECH2021}
\usepackage{xcolor, soul, multirow}
\usepackage{algorithm, algorithmic, graphicx}
\usepackage{listings}
\usepackage{hyperref}
\usepackage{fancyvrb}
\usepackage{url}

\title{NeMo Inverse Text Normalization: From Development To Production}
\name{Yang Zhang$^1$, Evelina Bakhturina$^1$, Kyle Gorman$^{2}$, Boris Ginsburg$^1$
\thanks{Preprint. Submitted to INTERSPEECH-21}
}

\address{
  $^1$NVIDIA \\
  $^2$Google and The Graduate Center, City University of New York}
\email{$^1$\{yangzhang, ebakhturina, bginsburg\}@nvidia.com $^2$kbg@google.com}

\begin{document}

\maketitle
\begin{abstract}
Inverse text normalization (ITN) converts spoken-domain automatic speech recognition (ASR) output into written-domain text to  improve the readability of the ASR output. Many state-of-the-art ITN systems use hand-written weighted finite-state transducer (WFST) grammars since this task has extremely low tolerance to unrecoverable errors. We introduce an open-source Python WFST-based library for ITN which enables a seamless path from development to production. We describe the specification of ITN grammar rules for English, but the library can be adapted for other languages. It can also be used for written-to-spoken text normalization. We  evaluate the NeMo ITN library using a modified version of the Google Text normalization dataset.
\end{abstract}

\section{Introduction}
\begin{figure*}[t]
  \centering
  \includegraphics[width=0.65\linewidth]{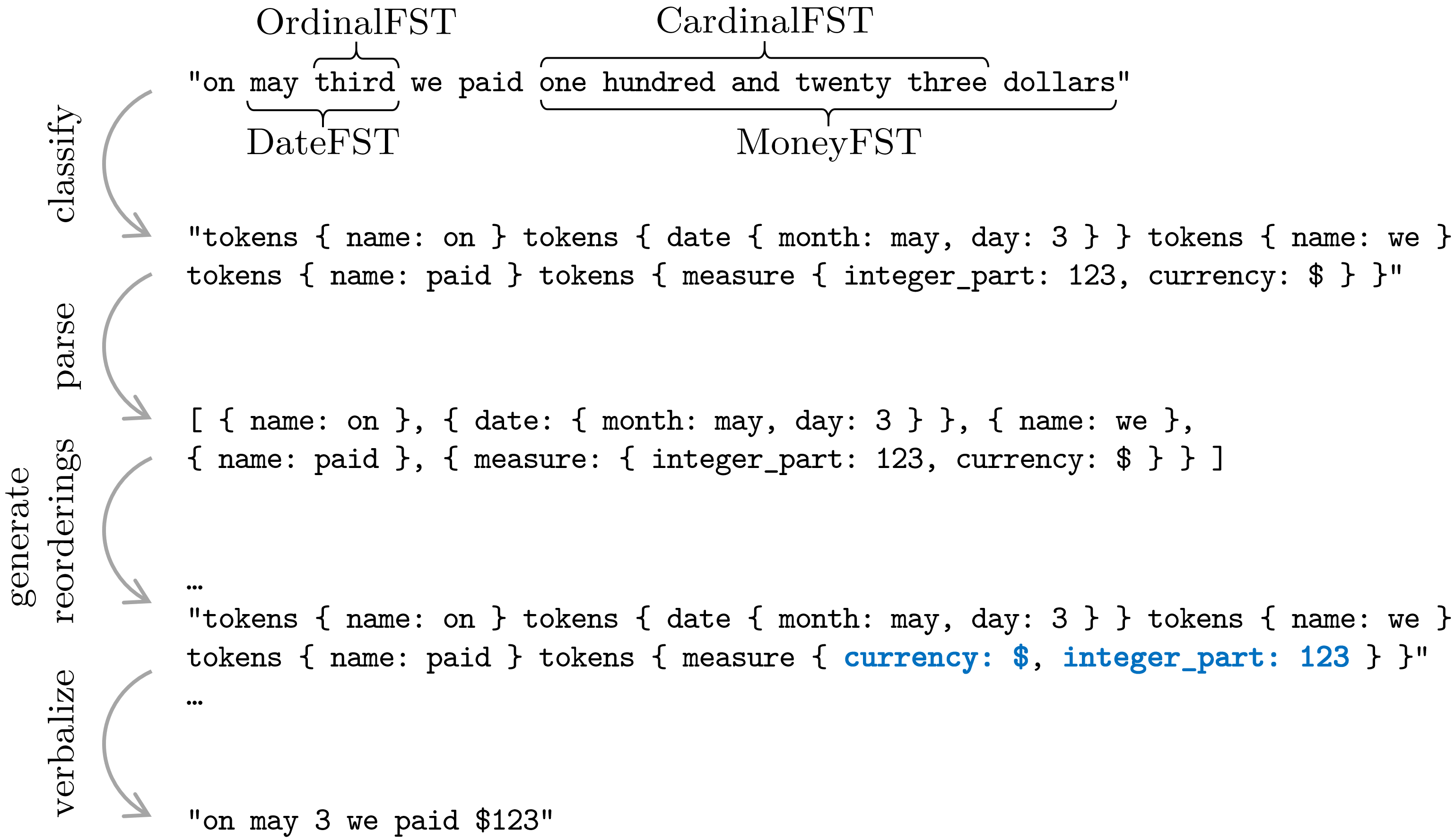}
  \caption{Pipeline of NeMo inverse text normalization}
  \label{fig:pipeline}
\end{figure*}

Inverse Text Normalization (ITN) is the process of converting spoken text to its written form. ITN is commonly used to convert the output of an automatic speech recognition (ASR) system to increase the readability for users and automatic downstream processes~\cite{sunkara2021neural}, such as extraction of time-related information. 
A sentence can be tokenized into plain tokens and non-standard words~\cite{taylor_tts}, also known as semiotic classes, e.g. CARDINALS, ORDINALS, DATE, see Figure \ref{fig:pipeline}.

ITN is one of the most challenging  natural language processing tasks: 1) it is highly language-dependent and especially tricky in inflected languages, 2) writing a complete set of grammars for ITN requires a lot of linguistic knowledge, and 3) data in its spoken form is scarce since it is rarely found on the web~\cite{5700892, ebden_sproat_2015}. 
ITN is heavily used in mobile spoken assistants  and demands very low error rates~\cite{coli_a_00349}. Further, some errors can be 
catastrophic if they alter the semantics of the input, e.g. \( \textit{``one hundred and twenty three dollars"} \rightarrow \textit{\$23}\), see \cite{coli_a_00349}.
Low tolerance towards unrecoverable errors is the main reason why most ITN systems in production are still largely rule-based, using weighted finite-state transducers (WFST)~\cite{Mohri2009, ebden_sproat_2015}. 
Recently, some  hybrid systems  have appeared which combine neural networks (NN)  with lightweight grammars, so-called text covering grammars~\cite{Sproat2017}. This can limit but not avoid unrecoverable errors, in cases where a NN's output is taken when no appropriate grammar is found~\cite{coli_a_00349}.

In this paper we introduce the NeMo ITN tool,
a Python package for ITN using WFST grammars. The framework is a Python extension of  \textit{Sparrowhawk} -- an open source version of \textit{Kestrel}~\cite{sparrowhawk}.  To write and compile grammars into WFSTs, the package uses  \textit{Pynini}~\cite{gorman-2016-pynini}, a Python toolkit built on top of \textit{OpenFst}. Its performance is much better than using FST classes implemented in Python.
While the Python framework offers great flexibility and ease of use for research, all created grammars can be exported and seamlessly integrated into \textit{Sparrowhawk} for production, see Figure \ref{fig:deployment}.
Currently, the package supports English only, but the toolkit is easily extendable due to its modular design and it can be adapted to other languages and tasks like text normalization. The Python environment enables easy combination of text covering grammars with NNs. 
Since a public dataset for ITN is not available, we adapted the Google Text Normalization dataset~\cite{Sproat2017} for ITN.

\section{Related work}

ITN is closely related to text normalization (TN), where the goal is to convert text from written form to spoken form. The ITN and TN systems can be classified into three types: 
\begin{enumerate}
    \item Rule-based systems, e.g. based on WFST grammars~\cite{ebden_sproat_2015}.
    \item NN-based models, typically using seq2seq models~\cite{large-context,mansfield-etal-2019-neural,PRAMANIK201915}.
    \item Hybrid systems that employ a weak covering grammar to filter and correct misreadings~\cite{formatting, ranking, sunkara2021neural, coli_a_00349, Sproat2017}.
\end{enumerate}
Most production ASR/TTS systems use text normalization systems   based on WFST grammars such as  \textit{Kestrel}~\cite{ebden_sproat_2015}. These systems consist of a huge set of language specific grammars. They are highly accurate. For example, \textit{Kestrel} has an accuracy of 99.9$\%$ on the Google TN test dataset~\cite{coli_a_00349}, whereas Apple's rule-based ITN has an estimated accuracy of 99$\%$ on internal data from an English virtual assistant application~\cite{apple}.
But it is hard to scale such a system across languages and the development takes significant effort and time, even for experienced linguists~\cite{coli_a_00349}.

The second type of TN systems are NN-based models, typically seq2seq ~\cite{large-context,mansfield-etal-2019-neural,PRAMANIK201915}.
Training such models requires large datasets. Collecting them manually shifts the challenge towards data labeling which also requires linguistic knowledge. Thus, datasets are usually generated using WFST-based systems~\cite{Sproat2017, coli_a_00349, sunkara2021neural,apple}.
Adding a new rule or new semiotic class to an NN-based model requires both new data and model retraining. But the biggest weakness is that these systems can make unrecoverable errors that are linguistically coherent but not information-preserving. 

The third type are hybrid systems that employ a weak covering
grammar to filter and correct the misreading~\cite{formatting, ranking, sunkara2021neural, coli_a_00349, Sproat2017}. These grammars are much smaller than those in a fully fledged rule-based system. However, in case of a missing grammar these systems fall back to using the neural model which is why these systems' unrecoverable errors are still very much decided by the coverage of existing grammars. 

There are not many public TN datasets. The only relatively large public TN dataset is the Google Text Normalization dataset~\cite{Sproat2017}, which was semi-automatically generated with \textit{Kestrel}~\cite{ebden_sproat_2015}. Google has also released the C++ framework \textit{Sparrowhawk}~\cite{sparrowhawk}, a pared-down open source version of \textit{Kestrel} without its grammars. 
The most popular library for WFST is \textit{OpenFst}, which is highly optimized for large graphs and inference. \textit{Pynini}~\cite{gorman-2016-pynini} is a popular Python package for writing WFST-based grammars with \textit{OpenFst} as a backend and can export compiled grammars into \textit{OpenFst} finite-state archive (FAR) files. 

\section{NeMo ITN framework}

\begin{figure*}[t]
  \centering
  \includegraphics[width=0.7\linewidth]{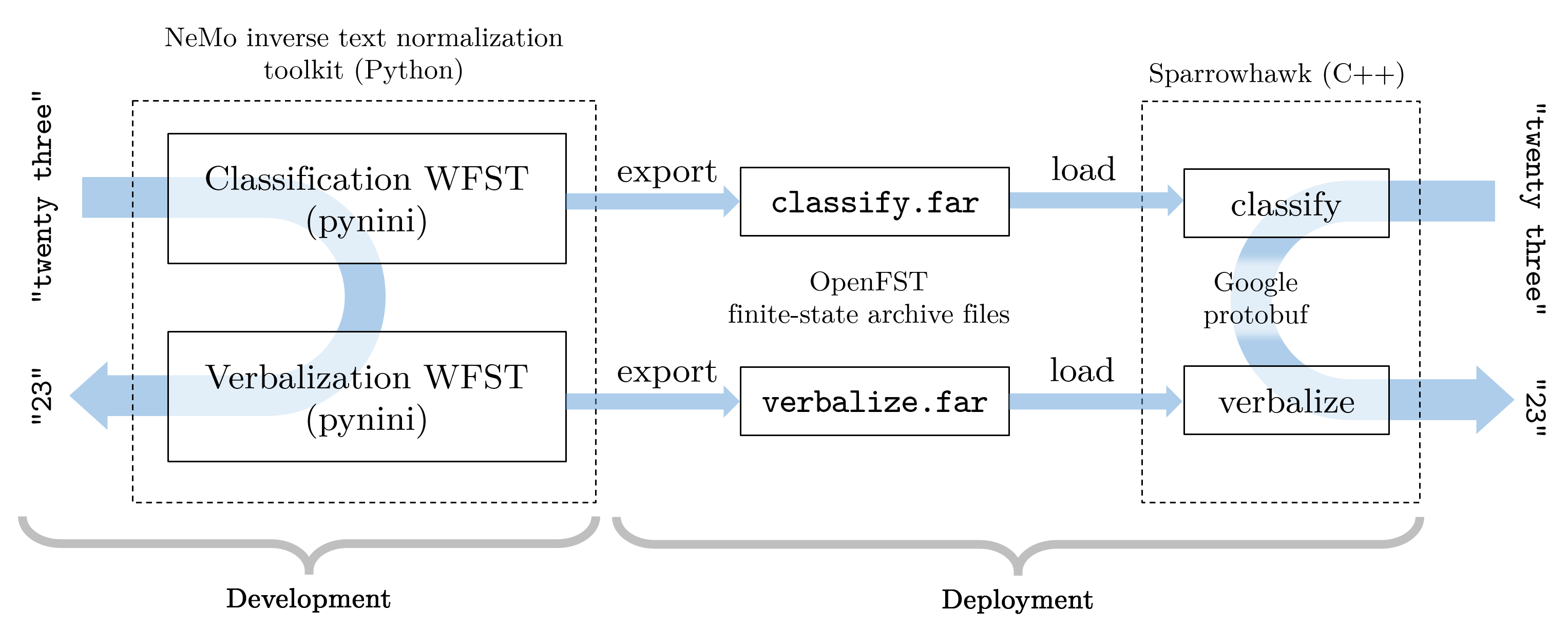}
  \caption{Schematic diagram of NeMo inverse text normalization development and deployment.}
  \label{fig:deployment}
\end{figure*}

Our goal was to build a Python extension of \textit{Sparrowhawk}~\cite{sparrowhawk}.  Following \textit{Sparrowhawk} design we use a two stage normalization pipeline that first detects semiotic tokens (classification) and then converts these to written form (verbalization). Both stages consume a single WFST grammar. NeMo ITN allows exporting these grammars as FAR files, which can then be directly plugged into \textit{Sparrowhawk} for production, see Figure \ref{fig:deployment}.

\subsection{API design}
The framework defines the following APIs that are called in sequence, see Figure \ref{fig:pipeline}:
\begin{lstlisting}
  classify: str -> str
  parse: str -> List[dict]
  generate_reorderings: List[dict] -> List[str]
  verbalize: str -> str
\end{lstlisting}



\textit{classify()} creates a linear automaton from the input string and composes it with the final classification WFST, which transduces numbers and inserts semantic tags.  

\textit{parse()} parses the tagged string into a list of key-value items representing the different semiotic tokens.

\textit{generate\_reorderings()} is a generator function which takes the parsed tokens and generates string serializations with different reorderings of the key-value items. This is important since WFSTs can only process input linearly, but the word order can change from spoken to written form (e.g., \( \textit{``three dollars"} \rightarrow \textit{\$3} \)). 
To align with \textit{Sparrowhawk}, we also have the option of fixing the order in case the verbalization should depend on the input, using the field \textit{preserve\_order: true}.
This is used for DATE, where \( \textit{``may third"} \rightarrow \textit{``may 3"}\) and \( \textit{``the third of may"} \rightarrow \textit{``3 may"}\).

\textit{verbalize()} takes the intermediate string representation and composes it with the final verbalization WFST, which removes the tags and returns the written form.  

\subsection{Pynini grammars}

We use the Python package \textit{Pynini}~\cite{gorman-2016-pynini} to write and compile grammars that transduce a string into another string. 
For all WFST, we use tropical semi-rings  -- the default WFST type in \textit{OpenFst} and \textit{Pynini}. This type is recommended for operations such as $shortestpath$ on the lattice for transduction, where the path weight is simply the summation of all arc weights along the path. 

We use the same taxonomy of semiotic classes as Sproat and Jaitly~\cite{Sproat2017}, see Table \ref{tab:result_test}.  
We define a Python class for each semiotic class, filled with grammars for that class. Each grammar class is solely responsible for its respective token inputs. Together with operations like union and Kleene closure they constitute the final WFSTs $ClassifyFst$ and $VerbalizeFst$ which can transduce an entire utterance with multiple tokens, see Figure \ref{fig:pipeline}.
Since this rule-based system requires a rule for every possible input we define $WordFst$ as the class to catch all PLAIN tokens. 
We added the class $WhiteListFst$ that transduces a token based on a lookup table that is predefined in a TSV file. In case of ambiguity, the white-list has always higher priority over other classes.

The most important class is $CardinalFst$, which required approximately 100 lines of code for English. Other grammars are built on top of it and usually contain only 10 to 20 lines of code. 
For $CardinalFst$ we define a minimal set of number mappings for $digits$, $teens$ and $ties$ and use  $pynini.string\_file$ to build a transducer that is the union of several string-to-string transductions from a TSV file. The rest of the graph is built from these building blocks by first creating a sub graph that consumes all three digit numbers. This is then composed with other graphs that consume quantities like $thousand$, $million$, $billion$, and so forth.   

\subsection{Grammar rule extension}
To add a new rule to an existing class, for example to support an additional date format, we can extend $DateFst$ by using \textit{Pynini} primitives.
To add an entirely new semiotic class, we create a classification and verbalization grammar class that inherits from the super class $GraphFst$. We append these to the final grammars $ClassifyFst$ and $VerbalizeFst$, respectively. 
Given a CARDINAL grammar, Figure \ref{fig:decimal_classify} shows an example of how to create a DECIMAL grammar for classification.

\begin{figure*}
    \centering
    \begin{lstlisting}[frame=single]
class DecimalFst(GraphFst):
  def __init__(self):
    super().__init__(name="decimal", kind="classify")
    fractional = pynini.string_file(get_abs_path("data/numbers/digit.tsv"))
    fractional |= pynini.string_file(get_abs_path("data/numbers/zero.tsv")) 
    fractional |= pynini.cross("o", "0")
    fractional = pynini.closure(fractional + delete_space) + fractional
    delete_point = pynutil.delete("point")
    tagged_fractional = pynutil.insert("fractional_part: \"") + fractional + pynutil.insert("\"")
    tagged_integer = pynutil.insert("integer_part: \"") + cardinal + pynutil.insert("\"")
    decimal = tagged_integer + delete_point + delete_extra_space + tagged_fractional
    self.fst = self.add_tokens(decimal).optimize()
    \end{lstlisting}
    \caption{Decimal grammar for classification}
    \label{fig:decimal_classify}
\end{figure*}
This will transduce \textit{``two point o five"} to the following:
\begin{lstlisting}
decimal { integer_part: "2" 
fractional_part: "05"}
\end{lstlisting}
The corresponding verbalization grammar is shown in Figure \ref{fig:decimal_verbalize}, which creates the final written form \textit{``2.05"}.
\begin{figure*}
    \centering
    \begin{lstlisting}[frame=single]
class DecimalFst(GraphFst):
  def __init__(self):
    super().__init__(name="decimal", kind="verbalize")
    integer = pynutil.delete("integer_part:") + delete_space
            + pynutil.delete("\"") + pynini.closure(NEMO_NOT_QUOTE, 1) + pynutil.delete("\"")
    fractional = pynutil.insert(".") + pynutil.delete("fractional_part:") + delete_space
               + pynutil.delete("\"") + pynini.closure(NEMO_NOT_QUOTE, 1) + pynutil.delete("\"")
    graph = integer + fractional
    self.fst = self.delete_tokens(graph).optimize()
    \end{lstlisting}
    \caption{Decimal grammar for verbalization}
    \label{fig:decimal_verbalize}
\end{figure*}

\subsection{Weight setting}

Given an input, there is often more than one matching path in a WFST lattice. 
Setting correct weights is crucial for choosing the desired path. We select the shortest path in a tropical semiring.
For example, given the input \textit{``twenty three"} what makes the grammar choose \textit{``23"} instead of \textit{``20 3"}? By default, all arcs in a path have weight 0, so it would be up to chance which path is chosen.

\begin{figure}[t]
  \centering
  \includegraphics[width=\linewidth]{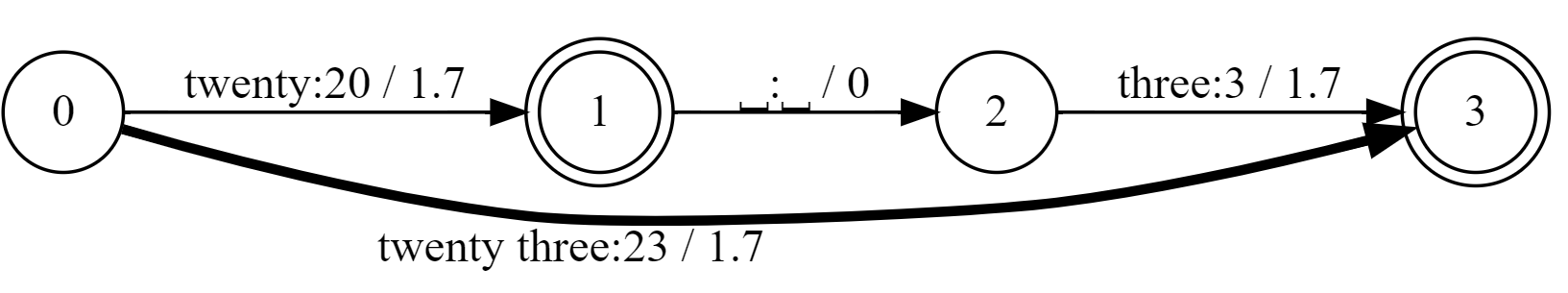}
  \caption{In case of ambiguity the path with the smallest sum of weights will be chosen. Here ``twenty three" is transduced to ``23" instead of ``20 3".}
  \label{fig:fst}
\end{figure}

We restrict every classification WFST apart from $WordFst$ to have a final weight $w \in (1, 2]$. 
This guarantees that in case of ambiguity, the shortest path is to extract the longest matching string instead of letting the same string be matched by two or more classes, since $w_1 + w_2 > 2 \geq w$. An example is shown in Figure \ref{fig:fst}.
We choose a large weight for $WordFst$ so the system will prefer matching an input to a semiotic class whenever possible.

\subsection{Deployment to production}
We use \textit{Sparrowhawk} as the production backend. 
The most important file for integration is $semiotic\_classes.proto$ which includes all predefined semiotic classes and their tags. This file needs to be adapted when adding new classes or tags. 
Another important file is $sparrowhawk\_configuration.ascii\_proto$ which is the entry point configuration.
It references $tokenizer.ascii\_proto$ and $verbalizer.ascii\_proto$ which contain the location of the classification and verbalization FAR files. We provide a script that automatically exports the WFST grammars to the dedicated location and runs \textit{Sparrowhawk} inside a docker container.

\subsection{Universal framework}
This framework is general enough to be applied to text normalization and other languages. 
In fact, \textit{Kestrel}~\cite{ebden_sproat_2015} and \textit{Sparrowhawk}~\cite{sparrowhawk} were originally intended for text normalization across a variety of languages.
To do this, we can replace the grammars for all classes. The classification grammars will be mostly language-specific, whereas the verbalization grammars should need fewer changes.

\section{Evaluation}

\subsection{Dataset}
The Google Text Normalization dataset~\cite{Sproat2017} consists of 1.1 billion words of English text from Wikipedia, divided across 100 files. The normalized text is obtained with the Kestrel text normalization system
~\cite{Sproat2017}.
Although a large fraction of this dataset can be reused for ITN by swapping input with output, the dataset is not bijective. For example,  \(\textit{1,000}\rightarrow \textit{``one thousand"}\), \(\textit{1000}\rightarrow \textit{``one thousand"}\), \(\textit{3:00pm}\rightarrow \textit{``three p m"}\), 
\(\textit{3 pm}\rightarrow \textit{``three p m"}\) are valid data samples for normalization but the inverse does not hold for ITN. As a consequence,  we expect the accuracy of ITN to be lower on the dataset (``orignal''), see Table \ref{tab:result_test}. Therefore, we used regex rules to disambiguate samples where possible (``cleaned'').

\subsection{Results}
We evaluated the ITN system on the original and cleaned dataset using the exact match score, the metric most commonly reported for TN ~\cite{Sproat2017,coli_a_00349}. 
The test results for semiotic tokens and full sentences are shown in Table \ref{tab:result_test}. Note that we only report results for semiotic classes for which we provided grammars. For some classes like DATE the samples are too heterogeneous so the accuracy is low.
For comparison, we used the \textit{word2number}\footnote{\url{https://pypi.org/project/word2number/}} Python package  that converts spoken numbers into written numbers. However, the \textit{word2number} only supports CARDINAL and ORDINAL and it cannot be applied to full utterances. It makes detrimental mistakes, for example \textit{``thirty million one hundred ninety thousand"} is falsely converted to ``30191190". In general, the exact match scores of \textit{word2number} are lower than those of NeMo ITN, with a exact match of 98.5\% for CARDINAL, and 78.65\% for DECIMAL on the cleaned dataset. 

The  other work that reported ITN results on the Google dataset is Sunkara et al~\cite{sunkara2021neural}, which uses a neural model with text covering grammars. Their best results of 0.9\% WER on full sentences on the test data is not comparable to the 12.7\% that we achieve with our tool, see Table \ref{tab:result_test}. However, they reported that their baseline FST model reaches 14.4\% WER which is similar to that of our tool. This shows that our tool provides a good baseline for text covering grammars with a neural network.

\begin{table}[thb]
\caption{\label{tab:result_test} Results on Google Text Normalization test dataset. We used 92K tokens of the final file. }
\centering
\resizebox{1.0\columnwidth}{!}{
\begin{tabular}{crrrrrr}
\toprule
  \multirow{2}{*}{\textbf{Class} }
 & \multicolumn{2}{c}{\textbf{Tokens}} 
 & \multicolumn{2}{c}{\textbf{Accuracy \%}} 
 & \multicolumn{2}{c}{\textbf{WER \%}}\\
 & \textbf{original} & \textbf{cleaned} & \textbf{original} & \textbf{cleaned}  & \textbf{original} & \textbf{cleaned}\\
   \hline
   SENTENCE & 6955 & 6955 & 65.56 & 73.73 & 12.7 & 10.14 \\ \hline
 PLAIN    & 62414 & 66284 & 99.1 & 99.1  & 1.1   & 1.1\\
 CARDINAL & 966   & 932   & 88.8 & 99.6  & 11.1  & 0.4\\
 ORDINAL  & 95    & 86    & 90.5 & 100.0 & 19.0  & 0.0\\
 DECIMAL  & 89    & 89    & 92.1 & 100.0 & 7.9   & 0.0\\
 MEASURE  & 127   & 127   & 26.0 & 96.1  & 151.4 & 3.0\\
 MONEY    & 29    & 29    & 37.9 & 93.1  & 67.8  & 5.7\\
 DATE     & 2607  & 2607  & 76.8 & 94.9  & 16.8  & 10.6\\
 \bottomrule
\end{tabular}}
\end{table}

\subsection{Failing cases}
Most failing cases come from  contextual disambiguation or the long tail of special cases.
For example, ``two pounds" can be either \textit{2 lb} or \textit{£2}, \textit{``second"} can be \textit{2nd} or \textit{s}.
A larger input context may distinguish between different classes, for example to detect ``£2" as MONEY if it is preceded by the word ``cost". This does not work for all cases, e.g. ``I paid two pounds". For this reason, when designing a complete system of grammars like \textit{Kestrel}, the number of grammars goes up exponentially. 
Other common failing cases are due to incomplete definitions. For example, if a measure acronym such as \textit{``volt"}$\rightarrow$\textit{v} is not predefined, the system will fail to transduce \textit{``two volt"}$\rightarrow$\textit{2 v}. These are relatively easy to add, but nonetheless need to be added explicitly.

\section{Conclusions}

We introduce the NeMo ITN toolkit -- an open source Python package for inverse text normalization based on WFSTs.  NeMo ITN is a Python extension for \textit{Sparrowhawk}, Google's open source C++ text normalization system. The framework provides a seamless route to \textit{Sparrowhawk} for deployment.  We define and evaluate a set of grammars for English using \textit{Pynini}, a toolkit built on top of \textit{OpenFst}. The toolkit is flexible and it can easily be adapted to other tasks such as text normalization and for text covering grammars in combination with neural models. We cleaned the Google Text Normalization dataset  and provided a baseline set of English grammars for inverse text normalization. 
The ITN package and the grammars are open-sourced in the NeMo toolkit~\cite{kuchaiev2019nemo}\footnote{ \url{https://github.com/NVIDIA/NeMo/tree/main/nemo_text_processing}}.

\section{Acknowledgements}
We would like to thank Richard Sproat for uploading the full Google Text Normalization dataset and for his help with  \textit{Sparrowhawk}. We would also like to thank Vitaly Lavrukhin, Chris Parisien, Elena Rastorgueva and our colleagues at NVIDIA for feedback.




\bibliographystyle{IEEEtran}
\bibliography{mybib}

\end{document}